\documentclass[11pt]{article}

\usepackage[final]{acl}

\usepackage{times}
\usepackage{latexsym}
\usepackage[T1]{fontenc}
\usepackage[utf8]{inputenc}
\usepackage{microtype}
\usepackage{inconsolata}
\usepackage{graphicx}
\usepackage{booktabs}
\usepackage{amsmath}
\usepackage{multirow}
\usepackage{tabularx}
\usepackage{adjustbox}

\title{A Subword Embedding Approach for Variation Detection in Luxembourgish User Comments}

\author{Anne-Marie Lutgen, Alistair Plum, Christoph Purschke \\
        University of Luxembourg, Esch-sur-Alzette, Luxembourg\\
        \texttt{\{anne-marie.lutgen, alistair.plum, christoph.purschke\}@uni.lu} \\}

\begin{document}
\maketitle
\begin{abstract}
This paper presents an embedding-based approach to detecting variation without relying on prior normalisation or predefined variant lists. The method trains subword embeddings on raw text and groups related forms through combined cosine and n-gram similarity. This allows spelling and morphological diversity to be examined and analysed as linguistic structure rather than treated as noise. Using a large corpus of Luxembourgish user comments, the approach uncovers extensive lexical and orthographic variation that aligns with patterns described in dialectal and sociolinguistic research. The induced families capture systematic correspondences and highlight areas of regional and stylistic differentiation. The procedure does not strictly require manual annotation, but does produce transparent clusters that support both quantitative and qualitative analysis. The results demonstrate that distributional modelling can reveal meaningful patterns of variation even in ``noisy'' or low-resource settings, offering a reproducible methodological framework for studying language variety in multilingual and small-language contexts.

\end{abstract}

\section{Introduction}
Variation in language is often treated as noise in NLP pipelines \cite{eisenstein-2013-bad, al-sharou-etal-2021-towards}. Spelling differences, orthographic inconsistencies, and regional forms are typically normalised or removed to simplify token space, which can erase sociolinguistic signal \citep{baron2008vard}. Work in sociolinguistics and large-corpus dialectology shows that such variation is systematic and informative for geography and social structure \citep{Grieve2019}. Subword modelling has long been used to handle non-standard forms in practice and improves classification in noisy settings \citep{Munro2010}.

For under-researched languages and varieties, identifying and extracting language variation remains challenging. Pre-processing tools such as VARD insert modern equivalents for historical spellings to aid search and tagging, relying on lexicons and edit-distance style methods, and target normalisation rather than discovery \citep{baron2008vard}. Embedding-based studies indicate that distributional models encode many types of spelling variation and near-orthographic similarity, though evaluations are typically based on curated sets of variant pairs and focus on representation quality \citep{Nguyen2020}. Closely related work in context-sensitive spelling correction uses word and character n-gram embeddings to map misspellings to canonical forms, again optimising correction and not mining new variants \citep{fivez2017unsupervised}. Research on dialectal change detection has brought models for geographic differences, but does not allow for directly extracting unconstrained orthographic families from raw text \citep{Jiang2020,Pham2024}. Broader multilingual analyses of surface-form overlap highlight that form-level variation carries structure that models can exploit \citep{Kallini2025}.

There are methods to represent variation and to correct it, and there are resources that label dialectal differences. What is less supported are methods that detect and mine previously unlisted variant families directly from raw text without seed lexicons, and that do so beyond strictly dialectal contrasts. We propose the methodology laid out in this paper to address this gap. With this methodology, we are able to discover candidate variation families from distributional evidence and provide transparent scores for downstream qualitative and quantitative analysis.

The main contributions of the research carried out and presented in this paper are:

\begin{itemize}
    \item[(1)] A reproducible methodology\footnote{\url{https://github.com/plumaj/vadamt}} for inducing lexical and orthographic variation from raw text using subword embeddings, similarity-based grouping, and controlled pruning without relying on predefined variant lists or normalisation rules.
    \item[(2)] A large-scale empirical study of variation in Luxembourgish user comments, showing that the automatically induced families capture systematic patterns and provide a structured basis for qualitative linguistic analysis. 
\end{itemize}

\section{Background}
\label{sec:background}
Luxembourgish is a small language situated in a dense multilingual environment, with extensive contact to both German and French. Its grammatical structure derives from Moselle Franconian \citep{Gilles2019}, while sustained contact with French has shaped its lexicon, borrowing patterns, and code-switching practices. Written Luxembourgish displays considerable orthographic and lexical diversity, especially in informal online settings, which makes it a challenge for NLP.

Initial work on computational methods for Luxembourgish was limited but established an initial foundation. \citet{adda-decker-etal-2008-developments} introduced the first tools and corpora for automatic processing. Subsequent studies examined characteristic orthographic phenomena \citep{snoeren-etal-2010-study} and provided early annotated resources for mixed-language processing \citep{lavergne-etal-2014-automatic}. 

Recently, research activity has increased noticeably. Work has expanded to sentiment analysis \citep{sirajzade-etal-2020-sentiment,Gierschek2022}, orthographic correction \citep{purschke2020attitudes}, syntactic annotation \citep{plum2024}, topic classification \citep{philippy2024}, comment moderation \citep{ranasinghe-etal-2023-publish}, and automatic normalisation \citep{lutgen2025}. A broader set of classification tasks, including named entity recognition, was provided by \citet{lothritz-etal-2022-luxembert}, and the generative benchmark \texttt{LuxGen} was introduced by \citet{plum2025}. These works illustrate the rapid growth of Luxembourgish NLP but also reveal gaps in coverage, consistency, and domain diversity.

Model development reflects a similar trajectory. Strategies range from cross-lingual transfer from German, as in \texttt{LuxGPT} \citep{bernardy2022}, to data augmentation with synthetic Luxembourgish text in \texttt{LuxemBERT} \citep{lothritz-etal-2022-luxembert}, and balanced multilingual pretraining for \texttt{LuxT5} \citep{plum2025}. Other models include \textsc{ENRICH4ALL} \citep{anastasiou-2022-enrich4all} for administrative-domain chatbots and the \textsc{LUX-ASR} speech recognition models \citep{gilles-etal-2023-asrlux,gilles-etal-2023-luxasr}. Together, these efforts demonstrate progress, yet available datasets remain fragmented and vary widely in size, annotation schemes, and linguistic phenomena.

One area that has received little explicit attention is lexical and orthographic variation. \citet{lutgen2025} develop a qualitative performance test to evaluate normalisation models for specific orthographic variants. In linguistics, the Variation Atlas by \citet{Gilles2021} represents the most comprehensive overview of phonological, lexical, grammatical and regional variants in Luxembourgish. This atlas is constructed by using an app to collect users' speech inputs for specific phenomena and socio-demographic data, which is then transcribed, analysed, and published \cite{schnessen}.

\section{Methodology}
\label{sec:method}
The methodology adopted in this study combines semi-supervised modelling with targeted qualitative analysis to identify lexical and orthographic variation directly from raw text, without relying on predefined dictionaries or normalisation rules. Throughout, spelling diversity is treated as a source of linguistic information rather than noise, allowing the unsupervised detection of previously unrecorded orthographic and mixed variants while ensuring transparency and reproducibility. This design supports large-scale induction alongside qualitative interpretation, and aligns with recent work arguing that normalisation can obscure meaningful patterns in non-standard and partly standardised varieties \citep{Grieve2019,Kallini2025}.

First, subword embeddings\footnote{In this work, we use the term subword embeddings to refer to embeddings constructed from fixed character n-grams, rather than learned segmentation-based subword vocabularies.} are trained on the raw corpus to obtain distributional representations that preserve orthographic detail. Second, these embeddings are used to induce groups of related forms through a combination of cosine and n-gram similarity, followed by controlled pruning and aggregation across relevant \textit{dimensions} such as users, time periods, or domains. Third, the automatically identified groups are examined manually to assess their linguistic coherence and to trace patterns that are not fully captured by numerical criteria. 

\subsection{Distributional Embeddings}
Before outlining the methodology in more detail, we briefly characterise what word- and subword-level embeddings encode. Distributional embeddings represent lexical items based on their patterns of co-occurrence in context, such that similarity in embedding space reflects shared semantic content, syntactic behaviour, and usage environments \citep{Turney2010,Levy2014}. Subword models extend this principle by incorporating character-level information, which allows orthographically related forms to be represented closely even when token frequencies are low or surface forms differ \citep{Bojanowski2017}. As a consequence, embedding similarity reflects semantic relatedness, morphosyntactic similarity, and orthographic overlap. This makes clustering in embedding space a suitable operation for identifying candidate groups of lexical variants, particularly in settings where variation manifests through both form and contextual usage \citep{Munro2010,Nguyen2020}.

\subsection{Stage 1: Training Subword Embeddings}
Embeddings are trained with FastText \cite{Bojanowski2017} using the configuration values \texttt{vector\_size}, \texttt{window}, \texttt{min\_count}, \texttt{epochs}, \texttt{min\_n}, \texttt{max\_n}, \texttt{sg}. These values are estimated in accordance with the size of the corpus, as well as with some testing of the variant families (as detected in the following stage).

The input is a JSON file containing a required \texttt{text\_field}. Optional fields specify the comparison dimension, such as \texttt{user\_id} or \texttt{date}. The corpus is streamed to manage memory, and basic token statistics are collected. Cleaning behaviour is minimal: Mentions beginning with \texttt{@} are removed before tokenisation, lowercasing is controlled by the \texttt{lowercase} flag.

\subsection{Stage 2: Identifying Variant Families}
After training, a candidate lexicon \(V\) is created from all tokens that meet the \texttt{min\_count} threshold. For each seed \(w \in V\), the method retrieves the top neighbours based on the values \texttt{open\_TOPN} or \texttt{strict\_TOPN}. Cosine similarity is computed as
\[
\cos(\mathbf{w}, \mathbf{v}) = \frac{\mathbf{w}\cdot\mathbf{v}}{\lVert\mathbf{w}\rVert\,\lVert\mathbf{v}\rVert}.
\]
Pairs are filtered according to the associated similarity threshold (\texttt{open\_TH} or \texttt{strict\_TH}). Then character n-gram Jaccard overlap is computed:
\[
J(w,v) = \frac{|G(w)\cap G(v)|}{|G(w)\cup G(v)|},
\]
where \(G(\cdot)\) contains all n-grams in the range \texttt{min\_n} to \texttt{max\_n}. Cosine and Jaccard values jointly determine whether two tokens belong to the same group.  

We implement two modes to help identify variant families. The \textit{open} mode forms a local star around each seed. The \textit{strict} mode builds an undirected graph and extracts connected components. Graph growth is limited by \texttt{DEGREE\_CAP}. Groups that do not reach \texttt{SNN\_MIN} members are removed. For the analysis presented in subsequent sections of this paper, we used \textit{strict} mode.

\paragraph{Scoring and Pruning}
For each group \(F\), we compute its size and the mean values of cosine similarity and Jaccard overlap. A cohesion score is the harmonic mean of these two averages. Groups are removed if they fail to reach the minimum size or if relative frequencies exceed the bound set by \texttt{MAX\_FREQ\_RATIO}. All pairwise scores are retained for inspection.

\paragraph{Dimension-Based Aggregation}
If a \texttt{dimension} field is provided, the method counts in how many distinct dimensions each variant appears and records the frequency of the most common dimension. For each variant we store its coverage, its top dimension, and the share that this dimension represents of its total frequency. These values feed into filters such as \texttt{MIN\_USERS} and \texttt{MAX\_FREQ\_RATIO}. The summary CSV lists these quantities for all groups.

\begin{table}[ht]
\centering
\small
\begin{tabular}{lcc}
\toprule
\textbf{Parameter} & \textbf{Key} & \textbf{Default} \\
\midrule
Lowercasing & \texttt{lowercase} & true \\
Comparison dimension & \texttt{dimension} & user\_id \\
Vector size & \texttt{vector\_size} & 100 \\
Context window & \texttt{window} & 5 \\
Min frequency & \texttt{min\_count} & 10 \\
Epochs & \texttt{epochs} & 10 \\
Skip-gram model & \texttt{sg} & 1 \\
Character n-gram range & \texttt{min\_n--max\_n} & 3--7 \\
Neighbours/seed (open) & \texttt{open\_TOPN} & 30 \\
Similarity thr. (open) & \texttt{open\_TH} & 0.75 \\
Neighbours/seed (strict) & \texttt{strict\_TOPN} & 100 \\
Similarity thr. (strict) & \texttt{strict\_TH} & 0.73 \\
Min family size & \texttt{SNN\_MIN} & 2 \\
Degree cap & \texttt{DEGREE\_CAP} & 200 \\
Min token length & \texttt{MIN\_LEN} & 3 \\
Min users per variant & \texttt{MIN\_USERS} & 3 \\
Max frequency ratio & \texttt{MAX\_FREQ\_RATIO} & 25 \\
\bottomrule
\end{tabular}
\caption{Main configuration parameters.}
\label{tab:params}
\end{table}

\paragraph{Configuration and Output}
Table \ref{tab:params} presents an overview of the parameters used and their defaults used for the purposes of this study. The method iterates through the vocabulary, computes cosine and Jaccard scores where needed, and constructs the final groups. The output consists of a JSONL file containing all groups with their members and a CSV summary with the main statistics. As this is an experimental study, the configuration of the values is based mainly on trial and error, by checking the variant families manually after each run. In contrast to normalisation tools such as VARD \citep{baron2008vard}, the procedure retains surface forms and measures their similarity instead of mapping them to canonical variants.

\subsection{Stage 3: Qualitative Analysis}
The model outputs variant families with their cosine and Jaccard values and the frequency of each member of the group. The first step in the qualitative analysis is to manually go over the families and approach the analysis with a bottom-up method. Based on the chosen dimension (i.e. user, time, etc.), the families represent for instance user variants in a similar context or variants over time in a similar context. Then, adopting a bottom-up approach, classifying the families is a straight-forward way to analyse the families based on the type of variation (orthographic, morphological, lexical, stylistic, regional, etc.). The families could also be semantically or functionally related, or the relation is not identifiable, which could be a category in itself. With the help of categorisation, the identification of patterns in the data is more feasible. 

\section{Luxembourgish User Comments}
We demonstrate the use of our methodology in the context of Luxembourgish user comments. The comments are part of the online media platform RTL\footnote{\url{https://rtl.lu}}, the main news broadcaster in Luxembourg and the only news broadcaster completely in Luxembourgish. The comments span from 2008 to 2024 and total roughly 1,42 million comments. As the use of Luxembourgish has been expanding in the written domain in the past 25 years and formal grammar teaching in school is still not properly regulated, we observe a high amount of variation in written texts. This is especially visible in informal domains, like online user comments. An in-depth analysis of variation in this domain in Luxembourgish has not been conducted yet. However, with our methodology, we can analyse a wide number of comments and classify the occurring variation. After applying stage 1 and 2, using the users as the comparative dimension, we start with the qualitative analysis. Using a bottom-up approach, we classify the families into 7 distinct categories, which are depicted in Table \ref{tab:cat}. As the families have a different number of words and also different phenomena appearing in one family, we decided for a multi-label approach. One family can have up to 3 categories. We illustrate the frequency of each category in Table \ref{tab:cat}. In the following, we describe each category and highlight findings related to language variation in Luxembourgish. 

\begin{table}[ht]
\centering
\small
\begin{tabular}{lc}
\toprule
\textbf{Category} & \textbf{Frequency} \\
\midrule
Orthographic &  394\\
Morphological &  222\\
Lexical &  115\\
Collocation &  21\\
Tokenisation &  14\\
Regional &  8\\
Other &  242\\
\bottomrule
\end{tabular}
\caption{Categories and frequency}
\label{tab:cat}
\end{table}

\newcounter{rownum}
\setcounter{rownum}{0}

\begin{table*}[ht!]
\begin{adjustbox}{max width=\textwidth, center}
\begin{tabular}{@{}llll@{}}
\toprule
\textbf{\#} & \textbf{Family} & \textbf{English}  & \textbf{Standard}\\ 
\midrule

(\stepcounter{rownum}\arabic{rownum})
    & Zäit, Zeit, Zait, Zéit, Zaït 
    & time 
    & Zäit \\

(\stepcounter{rownum}\arabic{rownum}) 
    & mir, mer, mier, mär, maer, miir, mäer 
    & we 
    & mir \\

(\stepcounter{rownum}\arabic{rownum}) 
    & mat, matt, maat 
    & with
    & mat \\

(\stepcounter{rownum}\arabic{rownum}) 
    & mecht, mécht, mescht, mëcht, mëscht
    & to do
    & mécht \\

(\stepcounter{rownum}\arabic{rownum}) 
    & wäit, weit, wait, wéit
    & far
    & wäit \\

(\stepcounter{rownum}\arabic{rownum}) 
    & sech, sëch, séch 
    & himself
    & sech \\

(\stepcounter{rownum}\arabic{rownum}) 
    & Numm, Num 
    & name
    & Numm \\

(\stepcounter{rownum}\arabic{rownum}) 
    & laang, lang
    & long
    & laang \\
    
(\stepcounter{rownum}\arabic{rownum}) 
    & Fehler, Feeler
    & mistake
    & Feeler \\

\bottomrule
\end{tabular}
\end{adjustbox}
\caption{Families in the category orthographic variation (One lexeme, different variants). }
\label{tab:variants}
\end{table*}

\paragraph{Orthographic} The orthographic category describes spellings that are not part of the official orthography \cite{ZLS19}. This includes the use of different graphemes to express the same word, which are often based on the word’s phonological properties. Since Luxembourgish has a high phoneme-grapheme correspondence in addition to an ideology that you can write how you speak, the language presents a wide range of orthographic variation. One example is \textit{laang} (lb. \textit{long}) where one family includes the orthographic variant \textit{lang}. The single vowel violates the quantity rule in the orthography \cite{ZLS19}, as a long vowel is written doubly when more than one consonant follows. However, since the orthography is not well known, both variants appear frequently. Additionally, this category also includes families that encompass different lexemes that are spelled incorrectly. One example for this case is \textit{krng, srng, dng, êng, öng} (lb. \textit{none, his, yours, one, one}). The correct spelling is \textit{keng, seng, deng, eng, eng}. In this instance, we can see two different spellings for \textit{eng} which has two different sources. The first one \textit{êng} represents more of a typical misspelling in comparison to \textit{öng} which represents a phonological variant expressed with the choice of <ö>. The phonological variant of \textit{eng} is pronounced with a rounded vowel which is then written as <ö> by some authors. This is still classified as orthographic variation since it violates the official orthography \cite{ZLS19}.

\paragraph{Morphological} The morphological category classifies all morphological variation. This encompasses conjugated verb forms, and inflections of nouns and adjectives in case, number, and gender. This also includes compounding nouns, clippings and conversions. One example is the family \textit{fillen, fillt, fille} (lb. \textit{to feel}) which includes the conjugated form \textit{fillt} and the deletion of the final <-n> before specific characters, known as the n-rule, in \textit{fille}.

\paragraph{Lexical} This category includes different lexemes that are semantically related. This includes synonyms and antonyms like the family \textit{méi, manner} (lb. \textit{more, less}) and lexical variants like \textit{dass, datt} (lb. \textit{that}). 

\paragraph{Collocation} The collocation category describes families with lexical items that form conventionalised combinations in daily use. For this category, the distinct words in a family form a collocation together. For instance, \textit{Gott, säi, Dank}\footnote{For readability, nouns in families are capitalized.} (lb. \textit{god, be, thanks}) forms one family of distinct items that frequently appear in the same context. Together, they form the collocation \textit{Gott säi Dank} meaning \textit{thank god}. 

\paragraph{Tokenisation} The tokenisation category describes families where the same word appears twice, but in one instance without the definite article \textit{d'} (lb. \textit{the}) attached to it and once with it attached to the word. One example for this is the family \textit{Zukunft, d'Zukunft} (lb. \textit{the future}). 

\paragraph{Regional} The regional category encompasses categories where regional varieties are visible in the graphemic representation. This category overlaps with the orthographic category, as it violates the official orthography \cite{ZLS19}. However, in this case the regional influence is visible and can be verified in the official Luxembourgish Variation Atlas \cite{Gilles2021}. 

\paragraph{Other} The other category includes families that have no distinct features or are not identifiable as meaningful words. This includes pragmatic expressions that are often used in online discourse like \textit{bla, blabla, ehm, hoho, tzzz} or non-identifiable words like \textit{fisk, ragnax, har, sed}.

\subsection{Orthographic Variation: Insights}

\begin{table*}[ht!]
\begin{adjustbox}{max width=\textwidth, center}
\begin{tabular}{@{}llll@{}}
\toprule
\textbf{\#} & \textbf{Family} & \textbf{English}  & \textbf{Standard}\\ 
\midrule

(\stepcounter{rownum}\arabic{rownum})
    & déi, wéi, méi, ewéi 
    & that, as, more, as 
    & déi, wéi, méi, ewéi \\

(\stepcounter{rownum}\arabic{rownum})
    & dèi, wèi, mèi
    & that, as, more 
    & déi, wéi, méi \\

(\stepcounter{rownum}\arabic{rownum}) 
    & dat, wat 
    & this, which 
    & dat, wat  \\

(\stepcounter{rownum}\arabic{rownum}) 
    & dad, wad 
    & this, which 
    & dat, wat  \\

(\stepcounter{rownum}\arabic{rownum}) 
    & weéi, deéi, eweéi, meéi
    & as, that, as, more
    & wéi, déi, ewéi, méi \\

(\stepcounter{rownum}\arabic{rownum}) 
    & méh, déh, wéh, ewéh
    & more, that, as, as
    & méi, déi, wéi, ewéi \\

\bottomrule
\end{tabular}
\end{adjustbox}
\caption{Families in the category orthographic variation (Different lexemes, one variant). }
\label{tab:variants_2}
\end{table*}

In this section, we present new insights from the orthographic category revealed by the clustering method. We examine several representative examples in more detail and discuss their impact.

\paragraph{One Lexeme, Different Variants} 
We start with examples, where one lexeme has several orthographic varieties in one single family. As the families are constructed based on similar neighbouring tokens in the embedding space, this grouping is straightforward. Interestingly, often not only variants of the same lexeme are part of the family but also the correctly spelled variant. This indicates that at least for these families, the lexemes of the orthographic Luxembourgish and the non-orthographic variants of Luxembourgish are aligned in the embedding space \cite{Cao2020Multilingual}. Additionally, this shows that the model is able to capture orthographic variants. In these cases, we either see multiple variants for one lexeme in one family, or only two variants, where one item is usually spelled correctly as indicated in Table \ref{tab:variants}. We observe that these categories include the most common orthographic variants in Luxembourgish. Especially in the families with only two variants, one variant is often the most common spelling variant.

\paragraph{Different Lexemes, One Variant}
The second insight for orthographic variation is the clustering of families of the same type of variation for different words across multiple families. For these families, we observe similar words, for instance \textit{dèi, wèi, mèi} (Table \ref{tab:variants_2} (11)) or \textit{dad, wad} (Table \ref{tab:variants_2} (13)) clustering together but with the same variant pattern in every word. A variant pattern is the same type of grapheme writing in different words, for example instead of the correct spelling of the diphthong <éi>, the graphemes <èi> are used consistently to represent the diphthong. Instances for these patterns are shown in Table \ref{tab:variants_2}. Additionally, not only variants cluster together but also the standard variants of these function words as shown in the instances (10) and (12) in Table \ref{tab:variants_2}. Since the instances are mostly function words, the clustering in itself is evident, however, the clustering of the identical variation pattern in different families indicates that these variants have social meaning which is represented in the embedding space. Therefore, the clustering shows that these words appear in similar topics or similar sentence structures that are linked via this variant. Another option would be a higher frequency for multiple authors writing about a similar topic. A more in-depth corpus analysis would give more insights into the social meaning of these variants. 

\subsection{Regional Variation: Insights}
Overall, our method only made 8 instances of regional variation visible. Due to an advanced state of dialect levelling in Luxembourgish, regional dialects have evolved into a national variety with some remaining lexical, phonological, and grammatical features \cite{Gilles1999}. Due to the informal nature of the comments, some phonological regional features are visible in the grapheme representation of the written lexeme. As these written forms differ from the official orthography, they also classify as orthographic variants. There are also instances of regional variants that are not orthographic variants but are part of the official dictionary for Luxembourgish\footnote{\url{https://lod.lu}} like \textit{mar} which is a regional variant of \textit{muer} (lb. \textit{tomorrow}). With the Variation Atlas \cite{Gilles2021} we can verify specific variants that are part of the 811 variant maps included in the atlas.

\begin{figure}[ht!]
\centering
\includegraphics[scale=0.36]{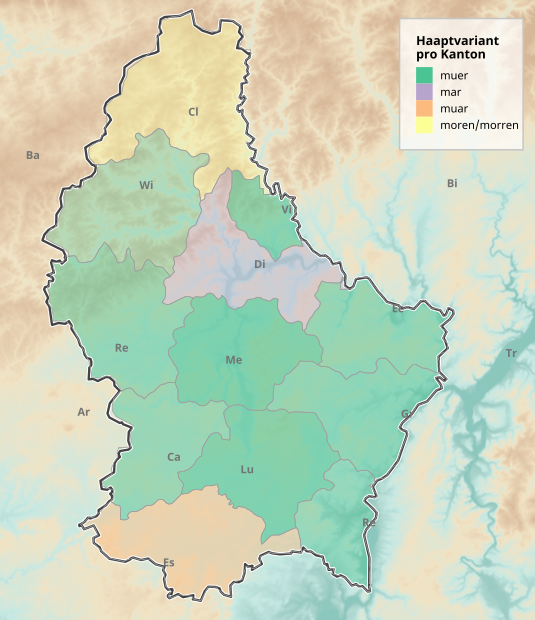}
\caption{Map of \textit{muer} variants \cite{Gilles2021} }
\label{fig:muer}
\end{figure}

One of those variants is the family \textit{muer, muar, moar} (lb. \textit{tomorrow}). Figure \ref{fig:muer} shows the map of the variant \textit{muer} and the regional variants \textit{mar, muar, moren}. Not shown in the map due to a low frequency of instances is the variant \textit{moar}, which is also part of the family in our analysis. The variant \textit{muar} is prevalent in the south of Luxembourg, whereas \textit{muer} is the most common variant in Luxembourg and considered the standard variant\footnote{Verified via the official dictionary for Luxembourgish.}. The variant \textit{moar} is mostly localised in the east of the country, but only a few participants (19) of the variation atlas survey have used that variant. However, in the comments 338 instances are recorded in comparison to 604 instances of \textit{muar}, and 6577 instances of \textit{muer}. This shows that \textit{moar} is still the least used variant in contrast to the most frequent variant \textit{muer}. However, it is still a common variant used in the comments, which was not clearly recorded before as the findings of the Variation Atlas did not indicate this. Additionally, the Jaccard values show a low overlap between authors using these variants, which indicates a consistent use of a variant instead of switching between variants in different contexts. 

\subsection{Lexical Variation: Insights}
The lexical variation category is one of the most heterogeneous categories in our analysis, as this category includes every family that encompasses different lexemes which are semantically related to each other. In addition to synonyms and antonyms, we also found common lexical variants in the data that are included in the Variation Atlas \cite{Gilles2021}. 

\begin{figure}[ht!]
\centering
\includegraphics[scale=0.4]{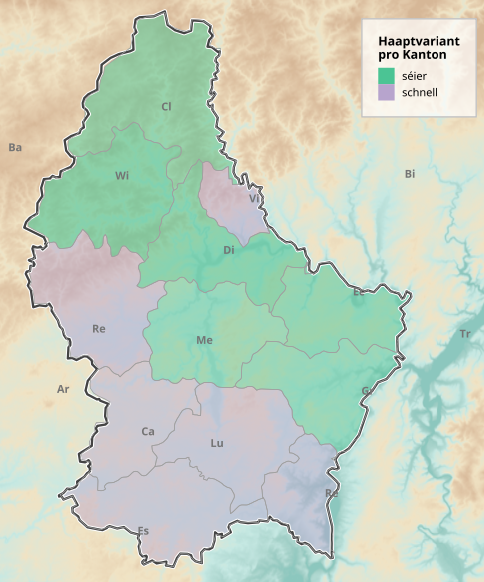}
\caption{Map of \textit{schnell} variants \cite{Gilles2021} }
\label{fig:schnell}
\end{figure}

Two interesting families are \textit{séier, schnell, seier} (lb. \textit{fast}) and \textit{säit, seit, sait, zanter, zenter, zënter, séit} (lb. \textit{since}). These families do not only show lexical variation, but also orthographic variation. In this section, we focus on the lexical side. Figure \ref{fig:schnell} illustrates the regional distribution of the use of \textit{séier} and \textit{schnell}. Overall, the frequency of use of both variants is nearly identical, at an almost equal split. The variant map illustrates some preference for the \textit{séier} variant in the north of the country, and some for \textit{schnell} in the south. However, the statistical analysis shows that only the north is significantly favouring the \textit{séier} variant \cite{Gilles2021}. One factor that influences the use of this variant significantly is age. The older the participants of the variation atlas survey are, the more the variant \textit{séier} is preferred. 

Similarly, we can observe these tendencies with \textit{säit}, \textit{zënter} and \textit{zanter}. \citet{Gilles2021} illustrates in the Variation Atlas that the variant \textit{säit} is overall favoured and regionally the most used variant. However, if we look at the factor age, the variant \textit{zënter} and \textit{zanter} are more popular with an older age group. It needs to be noted that \textit{zënter} and \textit{zanter} are phonological variants but belong the same lexical variant.   

Further, we also found families that are topically related. The family \textit{chat, chatgpt, gpt, chatbot} illustrates the more recent advances in AI and how these entered into the discourse in the user comments. Another example is the family \textit{Grexit, Frexit, Brexiteers, Lexit, Nexit, Luxexit} which encompasses word formations inspired by the expression \textit{Brexit}. By combining country names with the word \textit{exit} (e.g. \textit{France = Frexit}, \textit{Netherlands = Nexit}, \textit{Luxembourg = Luxexit}) we get a family of variants that is topically related to a withdrawal from the European Union for different countries. Interestingly, different countries cluster together, indicating a similar discourse on the topic for different countries.

\section{Discussion}

Previous work has demonstrated that specific lexical and orthographic variants tend to cluster in distributional space. Studies such as \citeposs{Nguyen2020} show that subword-based embeddings encode systematic spelling variation, and related work in dialectology and sociolinguistics has used clustering to analyse similarity between known varieties or regional forms. Crucially, much of this research starts from predefined units, such as known variants, dialect labels, or geographic groupings, and then examines how these cluster together. The analytical focus is typically on confirming that related forms or varieties occupy nearby positions in the embedding space.

The present study reverses this perspective. Instead of beginning with predefined variants, it detects and mines variants directly from co-occurrence and similarity patterns in the data. Clustering is thus used not as a validation tool, but as an exploratory mechanism that makes candidate forms visible without prior assumptions.

This methodological change is particularly relevant for Luxembourgish. Written Luxembourgish exhibits high orthographic diversity, partial standardisation, frequent borrowing, and code-switching, which make manual enumeration of variants difficult and incomplete. These properties create favourable conditions for unsupervised discovery of variation. This also applies to the dataset at the centre of our analysis. User comments provide dense, repetitive, and informal language use across many contributors, which supports distributional modelling of variation while retaining socially meaningful diversity.

Beyond identifying variant families, the approach opens up several directions for further analysis. One avenue concerns the role of different dimensions in shaping cluster structure. Future work will examine which types of variation cluster along which dimensions, for example whether certain families are associated with particular groups of users or whether historical and more recent forms of Luxembourgish can be distinguished through their distributional profiles. Exploring these questions will help clarify how different sources of variation interact and which analytical perspectives are best supported by this methodology.

The resulting clusters reveal patterns that are well aligned with linguistic intuition, but are not explicitly encoded in existing resources. This suggests that substantial lexical and orthographic variation remains undocumented. Resources such as the one analysed here therefore warrant further investigation, both to enrich descriptive accounts of Luxembourgish and to inform the development of computational models that better reflect actual language use.

Building on these considerations, we emphasise that the full codebase is released publicly to encourage reuse, scrutiny, and extension by other researchers. At the same time, our findings and limited further testing suggest that the proposed approach is not suited to highly standardised languages, where orthographic variation is limited and many relevant distinctions are already captured by existing lexical resources and normalisation pipelines. Its strengths instead lie in settings characterised by non-standardisation, activate variation, or incomplete codification, where spelling diversity encodes sociolinguistic and contextual information rather than noise. We therefore hope that this work enables and motivates applications to similar language varieties and research situations, including under-resourced or emerging standards. Beyond language-specific use cases, the method is also applicable to other domains and previously unexplored corpora, such as large web crawls or collections with limited metadata. In this sense, the contribution is less a universal solution for all languages than a transferable framework for studying variation in contexts where standard assumptions do not hold.

\section{Conclusion}
In this paper, we have made two main contributions. First, we describe a transparent and reproducible methodology for inducing lexical and orthographic variation directly from raw text, without relying on predefined variant lists or normalisation. Second, we present a large-scale empirical analysis of Luxembourgish user-generated text that documents variation as it is used in practice. Across seven analytically defined categories, the method identifies around 800 variant families, revealing systematic patterns of spelling and lexical diversity. The findings confirm earlier observations that related variants cluster in distributional space \citep{Nguyen2020}, while extending this insight by showing how such clusters can be mined directly from data rather than used only for validation.

Looking ahead, we plan to apply this methodology to additional Luxembourgish corpora in order to compare domains and writing contexts. This includes exploring whether user-level variation patterns can be characterised more systematically and assessing how well different available corpora reflect everyday language use. At the same time, initial experiments suggest that the approach is less effective for highly standardised languages, where orthographic variation is limited and distributional signals are weaker. This highlights that the method is particularly suited to languages and domains with active variation, and that its applicability depends on the sociolinguistic properties of the data.

\section*{Limitations} 
The findings in this study need to be interpreted with care. The corpus consists of user comments from a single online platform, which represents only a small portion of the Luxembourgish-speaking population. Patterns observed in this dataset therefore do not necessarily generalise to the wider speech community, nor do they capture the full range of regional, social, or stylistic variation present in Luxembourgish. The method identifies orthographic and lexical families based on distributional and subword similarity, which makes it sensitive to corpus composition and frequency effects. Rare variants may be missed, and high-frequency items can dominate neighbourhood structures. While the induced families provide useful candidates for analysis, their linguistic validity still depends on qualitative assessment. The results should thus be seen as a structured starting point for investigating Luxembourgish variation rather than a comprehensive account of the language.

A further limitation concerns the interpretation of the induced clusters. While the method identifies groups of closely related forms, it cannot by itself determine whether these patterns reflect linguistic variation, author-specific preferences, or temporal effects. In practice, these sources of variation are often intertwined in user-generated text, and distributional similarity alone does not allow them to be disentangled with certainty. Although dimension-based aggregation provides partial insight into how variants are distributed across users or time periods, the clustering process itself is agnostic to the underlying cause of similarity. As a result, the identified families should be interpreted as candidates for linguistic variation that require contextual and qualitative analysis to establish their nature.

\section*{Ethical Considerations}
This study uses publicly accessible user comments, but they remain sensitive textual data. All processing follows the terms of use of the platform from which the comments were collected. No attempt is made to identify individual users, and the analysis relies only on aggregated patterns such as variant frequencies and distribution across dimensions. Even though user identifiers are present in the raw data, they are treated only as categorical variables and are not (and could not be) interpreted as personal attributes. The dataset represents a self-selected set of online participants whose linguistic behaviour may differ from that of the wider population, and care should be taken not to attribute group-level characteristics to individual users. Finally, automatically induced variant families can reflect social or regional differences, but these patterns should be interpreted with caution to avoid reifying stereotypes or overgeneralising from limited data. The methodological framework is intended for linguistic analysis rather than profiling or prediction of individuals.

\section*{Acknowledgments}
This research was supported by the Luxembourg National Research Fund (Project code: C22/SC/117225699). 
The experiments reported in this paper were conducted on the MeluXina high-performance computing infrastructure, made available through an allocation granted by the University of Luxembourg on the EuroHPC supercomputer hosted by LuxProvide.
We would like to thank the TRAVOLTA project partners for their invaluable advice and guidance on this project.

\bibliography{custom}

@inproceedings{al-sharou-etal-2021-towards,
    title = "{Towards a Better Understanding of Noise in Natural Language Processing}",
    author = "Al Sharou, Khetam  and
      Li, Zhenhao  and
      Specia, Lucia",
    editor = "Mitkov, Ruslan  and
      Angelova, Galia",
    booktitle = "Proceedings of RANLP",
    month = sep,
    year = "2021",
    url = "https://aclanthology.org/2021.ranlp-1.7/",
}

@inproceedings{eisenstein-2013-bad,
    title = "What to do about bad language on the internet",
    author = "Eisenstein, Jacob",
    editor = "Vanderwende, Lucy  and
      Daum{\'e} III, Hal  and
      Kirchhoff, Katrin",
    booktitle = "Proceedings of NAACL-HLT",
    month = jun,
    year = "2013",
    url = "https://aclanthology.org/N13-1037/",
}

@article{schnessen,
author = {Entringer, Nathalie and Gilles, Peter and Martin, Sara and Purschke, Christoph},
year = {2021},
month = {01},
title = {Schnëssen. Surveying language dynamics in Luxembourgish with a mobile research app},
volume = {7},
journal = {Linguistics Vanguard},
doi = {10.1515/lingvan-2019-0031}
}

@book{Gilles1999,
  author    = {Gilles, Peter},
  title     = {Dialektausgleich im Lëtzebuergeschen: Zur phonetisch-phonologischen Fokussierung einer Nationalsprache},
  year      = {1999},
  address   = {Tübingen, Germany},
  publisher = {Niemeyer},
  url       = {https://hdl.handle.net/10993/4339}
}

@inproceedings{
Cao2020Multilingual,
title={Multilingual Alignment of Contextual Word Representations},
author={Steven Cao and Nikita Kitaev and Dan Klein},
booktitle={Proceedings of ICLR},
year={2020},
url={https://openreview.net/forum?id=r1xCMyBtPS}
}

@misc{Gilles2021,
  author       = {Gilles, Peter},
  title        = {Variatiounsatlas vum Lëtzebuergeschen},
  year         = {2021},
  howpublished = {\url{https://infolux.uni.lu/variatiounsatlas}},
  note         = {Accessed: 15.10.2025}
}

@book{ZLS19,
  title = {{D'L{\"e}tzebuerger Orthografie}},
  editor = {{Zenter fir d'L{\"e}tzebuerger Sprooch}},
  year = 2019,
  publisher = {Zenter fir d'L{\"e}tzebuerger Sprooch},
  address = {Stroossen},
  isbn = {978-99959-1-163-8},
  langid = {luxembourgish},
}

@article{baron2008vard,
  title={{VARD2: A tool for dealing with spelling variation in historical corpora}},
  author={Baron, Alistair and Rayson, Paul},
  year={2008}
  
}

@article{fivez2017unsupervised,
  title={{Unsupervised Context-Sensitive Spelling Correction of English and Dutch Clinical Free-Text with Word and Character N-Gram Embeddings}},
  author={Fivez, Pieter and {\v{S}}uster, Simon and Daelemans, Walter},
  journal={Computational Linguistics in the Netherlands Journal},
  year={2017}
}

@inproceedings{Munro2010,
	title = {Subword {Variation} in {Text} {Message} {Classification}},
	url = {https://aclanthology.org/N10-1075},
	urldate = {2024-12-17},
	booktitle = {Proceedings of NAACL-HLT},
	author = {Munro, Robert and Manning, Christopher D.},
	editor = {Kaplan, Ron and Burstein, Jill and Harper, Mary and Penn, Gerald},
	month = jun,
	year = {2010},
}

@inproceedings{Nguyen2020,
	title = {Do {Word} {Embeddings} {Capture} {Spelling} {Variation}?},
	url = {https://www.aclweb.org/anthology/2020.coling-main.75},
	doi = {10.18653/v1/2020.coling-main.75},
	language = {en},
	urldate = {2025-09-25},
	booktitle = {Proceedings of ICCL},
	author = {Nguyen, Dong and Grieve, Jack},
	year = {2020},
}

@misc{Kallini2025,
	title = {False {Friends} {Are} {Not} {Foes}: {Investigating} {Vocabulary} {Overlap} in {Multilingual} {Language} {Models}},
	shorttitle = {False {Friends} {Are} {Not} {Foes}},
	url = {http://arxiv.org/abs/2509.18750},
	doi = {10.48550/arXiv.2509.18750},
	urldate = {2025-09-30},
	publisher = {arXiv},
	author = {Kallini, Julie and Jurafsky, Dan and Potts, Christopher and Bartelds, Martijn},
	month = sep,
	year = {2025},
	note = {arXiv:2509.18750 [cs]},
}

@article{Grieve2019,
	title = {Mapping {Lexical} {Dialect} {Variation} in {British} {English} {Using} {Twitter}},
	volume = {2},
	issn = {2624-8212},
	url = {https://www.frontiersin.org/journals/artificial-intelligence/articles/10.3389/frai.2019.00011/full},
	doi = {10.3389/frai.2019.00011},
	language = {English},
	urldate = {2025-11-04},
	journal = {Frontiers in AI},
	author = {Grieve, Jack and Montgomery, Chris and Nini, Andrea and Murakami, Akira and Guo, Diansheng},
	month = jul,
	year = {2019},
}

@inproceedings{Pham2024,
	title = {Towards {Better} {Inclusivity}: {A} {Diverse} {Tweet} {Corpus} of {English} {Varieties}},
	shorttitle = {Towards {Better} {Inclusivity}},
	url = {https://aclanthology.org/2024.law-1.6/},
	urldate = {2025-11-04},
	booktitle = {Proceedings of {LAW} (ACL)},
	author = {Pham, Nhi and Pham, Lachlan and Meyers, Adam},
	editor = {Henning, Sophie and Stede, Manfred},
	month = mar,
	year = {2024},
}

@inproceedings{Jiang2020,
	title = {{DialectGram}: {Automatic} {Detection} of {Dialectal} {Changes} with {Multi}-geographic {Resolution} {Analysis}},
	shorttitle = {{DialectGram}},
	url = {https://aclanthology.org/2020.scil-1.18/},
	booktitle = {Proceedings of the {Society} for {Computation} in {Linguistics}},
	author = {Jiang, Hang and Hong, Haoshen and Chen, Yuxing and Kulkarni, Vivek},
	editor = {Ettinger, Allyson and Jarosz, Gaja and Pater, Joe},
	year = {2020},
}

@article{purschke2020attitudes,
	title        = {{Attitudes Toward Multilingualism in Luxembourg. A Comparative Analysis of Online News Comments and Crowdsourced Questionnaire Data}},
	author       = {Purschke, Christoph},
	year         = 2020,
	journal      = {Frontiers in AI},
	volume       = 3
}

@inproceedings{adda-decker-etal-2008-developments,
	title        = {{Developments of ``L{\"e}tzebuergesch{''} Resources for Automatic Speech Processing and Linguistic Studies}},
	author       = {Adda-Decker, Martine  and Pellegrini, Thomas  and Bilinski, Eric  and Adda, Gilles},
	year         = 2008,
	month        = {may},
	booktitle    = {{Proceedings of LREC}},
	url          = {http://www.lrec-conf.org/proceedings/lrec2008/pdf/855\%5Fpaper.pdf},
}

@phdthesis{Gierschek2022,
	title        = {{Detection of Sentiment in Luxembourgish User Comments}},
	author       = {Daniela Gierschek},
	year         = 2022,
	url          = {http://hdl.handle.net/10993/50533},
	school       = {University of Luxembourg}
}

@inproceedings{sirajzade-etal-2020-sentiment,
	title        = {{An Annotation Framework for Luxembourgish Sentiment Analysis}},
	author       = {Sirajzade, Joshgun and Gierschek, Daniela and Schommer, Christoph},
	year         = 2020,
	booktitle    = {{Proceedings of SLTU-CCURL (LREC)}},
}

@inproceedings{ranasinghe-etal-2023-publish,
	title        = {{Publish or Hold? {{Automatic}} Comment Moderation in {{Luxembourgish}} News Articles}},
	author       = {Ranasinghe, Tharindu and Plum, Alistair and Purschke, Christoph and Zampieri, Marcos},
	year         = 2023,
	month        = sep,
	booktitle    = {Proceedings of RANLP},
}

@mastersthesis{bernardy2022,
	title        = {{A Luxembourgish GPT-2 Approach Based on Transfer Learning}},
	author       = {Bernardy, Laura},
	year         = 2022,
	school       = {University of Trier}
}

@inproceedings{lavergne-etal-2014-automatic,
	title        = {{Automatic language identity tagging on word and sentence-level in multilingual text sources: a case-study on {L}uxembourgish}},
	author       = {Lavergne, Thomas  and Adda, Gilles  and Adda-Decker, Martine  and Lamel, Lori},
	year         = 2014,
	month        = may,
	booktitle    = {Proceedings of {LREC}},
	url          = {http://www.lrec-conf.org/proceedings/lrec2014/pdf/732\%5FPaper.pdf},
}

@inproceedings{snoeren-etal-2010-study,
	title        = {{The Study of Writing Variants in an Under-resourced Language: Some Evidence from Mobile N-Deletion in {L}uxembourgish}},
	author       = {Snoeren, Natalie D.  and Adda-Decker, Martine  and Adda, Gilles},
	year         = 2010,
	month        = may,
	booktitle    = {Proceedings of {LREC}},
	url          = {http://www.lrec-conf.org/proceedings/lrec2010/pdf/258\%5FPaper.pdf},
}

@inproceedings{philippy2024,
	title        = {{Forget {NLI}, Use a Dictionary: Zero-Shot Topic Classification for Low-Resource Languages with Application to {L}uxembourgish}},
	author       = {Philippy, Fred  and Haddadan, Shohreh  and Guo, Siwen},
	year         = 2024,
	month        = may,
	booktitle    = {Proceedings of SIGUL (LREC-COLING)},
	url          = {https://aclanthology.org/2024.sigul-1.13},
}

@incollection{Gilles2019,
	title        = {{39. Komplexe \"{U}berdachung II: Luxemburg. Die Genese Einer Neuen Nationalsprache}},
	author       = {Peter Gilles},
	year         = 2019,
	booktitle    = {{Sprache und Raum - Ein internationales Handbuch der Sprachvariation. Volume 4 Deutsch}},
	publisher    = {De Gruyter Mouton},
	address      = {Berlin, Boston},
	doi          = {doi:10.1515/9783110261295-039},
	isbn         = 9783110261295,
	url          = {https://doi.org/10.1515/9783110261295-039},
	editor       = {Joachim Herrgen and J\"{u}rgen Erich Schmidt}
}

@inproceedings{plum2024,
    title = "{L}ux{B}ank: The First {U}niversal {D}ependency Treebank for {L}uxembourgish",
    author = {Plum, Alistair  and
      D{\"o}hmer, Caroline  and
      Milano, Emilia  and
      Lutgen, Anne-Marie  and
      Purschke, Christoph},
    editor = {Dakota, Daniel  and
      Jablotschkin, Sarah  and
      K{\"u}bler, Sandra  and
      Zinsmeister, Heike},
    booktitle = "Proceedings of TLT",
    month = dec,
    year = "2024",
    url = "https://aclanthology.org/2024.tlt-1.4/",
}

@inproceedings{plum2025,
    title = "Text Generation Models for {L}uxembourgish with Limited Data: A Balanced Multilingual Strategy",
    author = "Plum, Alistair  and
      Ranasinghe, Tharindu  and
      Purschke, Christoph",
    editor = "Scherrer, Yves  and
      Jauhiainen, Tommi  and
      Ljube{\v{s}}i{\'c}, Nikola  and
      Nakov, Preslav  and
      Tiedemann, Jorg  and
      Zampieri, Marcos",
    booktitle = "Proceedings of VarDial (COLING)",
    month = jan,
    year = "2025",
    url = "https://aclanthology.org/2025.vardial-1.7/",
}

@inproceedings{lutgen2025,
    title = "Neural Text Normalization for {L}uxembourgish Using Real-Life Variation Data",
    author = "Lutgen, Anne-Marie  and
      Plum, Alistair  and
      Purschke, Christoph  and
      Plank, Barbara",
    editor = "Scherrer, Yves  and
      Jauhiainen, Tommi  and
      Ljube{\v{s}}i{\'c}, Nikola  and
      Nakov, Preslav  and
      Tiedemann, Jorg  and
      Zampieri, Marcos",
    booktitle = "Proceedings of VarDial (COLING)",
    month = jan,
    year = "2025",
    url = "https://aclanthology.org/2025.vardial-1.9/",
}

@inproceedings{lothritz-etal-2022-luxembert,
	title        = {{LuxemBERT: Simple and Practical Data Augmentation in Language Model Pre-Training for Luxembourgish}},
	author       = {Lothritz, Cedric  and Lebichot, Bertrand  and Allix, Kevin  and Veiber, Lisa  and Bissyande, Tegawende  and Klein, Jacques  and Boytsov, Andrey  and Lefebvre, Cl{\'e}ment  and Goujon, Anne},
	year         = 2022,
	month        = jun,
	booktitle    = {{Proceedings of LREC}},
	url          = {https://aclanthology.org/2022.lrec-1.543}
}

@inproceedings{anastasiou-2022-enrich4all,
	title        = {{ENRICH4ALL: A First Luxembourgish BERT Model for a Multilingual Chatbot}},
	author       = {Anastasiou, Dimitra},
	year         = 2022,
	month        = jun,
	booktitle    = {{Proceedings of SIGUL}},
	publisher    = {ELRA},
	url          = {https://aclanthology.org/2022.sigul-1.27}
}

@inproceedings{gilles-etal-2023-luxasr,
	title        = {{LUX-ASR: Building an ASR system for the Luxembourgish language}},
	author       = {Gilles, Peter and Hosseini Kivanani, Nina and Hillah, L\'{e}opold Edem Ayit\'{e}},
	year         = 2023,
	booktitle    = {Proceedings of IEEE Spoken Language Technology Workshop},
}

@inproceedings{gilles-etal-2023-asrlux,
	title        = {{ASRLUX: Automatic Speech Recognition for the Low-Resource Language Luxembourgish}},
	author       = {Gilles, Peter and Hillah, L\'{e}opold Edem Ayit\'{e} and Hosseini Kivanani, Nina},
	year         = 2023,
	booktitle    = {Proceedings of the International Congress of Phonetic Sciences},
}

@article{Turney2010,
  author  = {Turney, Peter D. and Pantel, Patrick},
  title   = {From Frequency to Meaning: Vector Space Models of Semantics},
  journal = {Journal of Artificial Intelligence Research},
  year    = {2010},
  volume  = {37},
  url     = {https://www.jair.org/index.php/jair/article/view/10640}
}

@inproceedings{Levy2014,
  author    = {Levy, Omer and Goldberg, Yoav},
  title     = {Neural Word Embedding as Implicit Matrix Factorization},
  booktitle = {Proceedings of NeurIPS},
  year      = {2014},
}

@inproceedings{Bojanowski2017,
  author    = {Bojanowski, Piotr and Grave, Edouard and Joulin, Armand and Mikolov, Tomas},
  title     = {Enriching Word Vectors with Subword Information},
  booktitle = {Transactions of ACL},
  year      = {2017},
  url       = {https://aclanthology.org/Q17-1010}
}



\end{document}